\documentclass[twoside,11pt]{article}

\usepackage{jmlr2e}
\usepackage{url}
\usepackage{listings}

\ShortHeadings{PyPLN}{Coelho et al.}
\firstpageno{1}

\begin{document}

\title{PyPLN: a Distributed Platform for Natural Language Processing}

\author{\name Fl\'avio Code\c{c}o Coelho \email fccoelho@fgv.br \\
       \addr School of Applied Mathematics\\
       Funda\c{c}\~ao Getulio Vargas\\
       Rio de Janeiro, RJ 22250-900, Brazil
       \AND
       \name Renato Rocha Souza \email renato.souza@fgv.br \\
       \addr School of Applied Mathematicsl\\
       Funda\c{c}\~ao Getulio Vargas\\
       Rio de Janeiro, RJ 22250-900, Brazil
       \AND
       \name \'Alvaro Justen \email alvaro.justen@fgv.br \\
        \addr School of Applied Mathematicsl\\
        Funda\c{c}\~ao Getulio Vargas\\
        Rio de Janeiro, RJ 22250-900, Brazil
       \AND
       \name Fl\'avio Amieiro \email flavioamieiro@gmail.com \\
              \addr School of Applied Mathematics \\
              Funda\c{c}\~ao Getulio Vargas\\
              Rio de Janeiro, RJ 22250-900, Brazil       
       \AND
       \name Heliana Mello \email hmello@ufmg.br \\
       \addr Faculdade de Letras\\
       Universidade Federal de Minas Gerais\\
        Belo Horizonte, MG 31270-901\\
       }

 \editor{}

\maketitle

\begin{abstract}
This paper presents PyPLN\footnote{\url{pypln.org}} a distributed platform for
Natural Language Processing.  PyPLN leverages a vast array of NLP and text
processing open
source tools, managing the distribution of the workload on a variety of
configurations: from a single server to a cluster of linux servers. PyPLN is
developed purely in Python but makes it very easy to incorporate other
softwares for specific tasks, as long as a linux version is available. PyPLN
facilitates analyses both at document and corpus levels, simplifying management
and publication of corpora and analytical results through an easy to use web
interface. In the current (beta) release, it supports English and Portuguese
languages (with support to other languages planned for future releases). To
handle texts in Portuguese language, PyPLN uses the PALAVRAS
parser\citep{Bick2000} if the user has it available\footnote{PALAVRAS must be
purchased and installed}.
Currently PyPLN offers the following features: Text extraction with encoding
normalization (to UTF-8), part-of-speech tagging, token frequency, semantic
annotation, n-gram extraction, 
word and sentence repertoire, and  full-text search across corpora. The platform
is licensed as GPL-v3.

\end{abstract}


\section{Introduction}

The PyPLN platform is a research project in active development in the School of
Applied Mathematics of Funda\c{c}\~ao Getulio Vargas, Brazil. Its main goal is
to make available a scalable computational platform for a variety of
language-based analyses. Its main target audience is the academic community,
where it can have a powerful impact by making sophisticated computational
analyses doable without the requirement of programming skills on the part of the
user.
Among the many features already available we can cite: Simplified
access to corpora with interactive visualization tools, text extraction from
HTML and PDF documents, Encoding detection and conversion to utf-8, Language
detection, Tokenization, full-text search across corpora, part-of-speech
tagging, Word and sentence level statistics, n-gram extraction and  word
concordance. Many more features are in development and should become available
soon, such as: semantic annotation, sentiment and text polarity
analysis, automatic social network information monitoring, stylistic analysis
and the generation of Knowledge Organization Systems such as ontologies and
thesauri. PyPLN aims for unrivaled ease of use, and wide availability, through
its web interface and full support to Portuguese language. Besides being a free,
uncomplicated research platform for language scholars capable of handling large
corpora, PyPLN is also a free software platform for distributed text processing,
which can be downloaded and installed by users on their own infrastructure.

The motivations for the development of PyPLN are multifaceted; however, its
major
inspiration comes from the current demands spanning from the myriad
possibilities
for textual-based analysis inspired by the ever growing number of
unstructured masses of textual data in digital form. A good example of this type
of data are the streams of human communication in social platforms such as
Twitter and Facebook; These social platforms alone have provoked an
extraordinary surge in the development  of natural language processing (NLP)
tools. But the applications of NLP are not limited to this area. Traditional
document collections are migrating to digital format and becoming a huge ``new''
market for NLP.

Despite the availability of a number of commercial and very good quality free
open source computational solutions for English (e.g GATE,  OpenNLP, UIMA), and
even complete programming frameworks for NLP (e.g NLTK for Python, NLP tasks for
R), Portuguese suffers from a scarcity of tools for the automatic treatment of
language. PyPLN tries to fill this niche by providing a solution which covers
both the needs of end-users (via the web interface), and the computer savvy
analyst (via the PyPLN API) while providing the computational resources in a
scalable platform capable of handling very large corpora.

PyPLN is free software licensed under GPLv3. The source code can be downloaded
from \url{pypln.org}.

\section{The Software}
\subsection{Architecture}
At the core of any text mining solution is the need to quickly put together
custom analytical workflows. Custom solutions are key, due to the
ample applicability of these tools: for every new data set there is a large set
of original analyses which can be proposed. Though the number of possible
analytical scenarios is very large, they can be framed by the recombination of
relatively simple smaller analytical steps. For example: no textual analysis can
be performed without some form of tokenization, for instance. These
intermediary results are calculated once for each document and cached for future
uses. The
modularity of these analytical steps leads to great performance gains by
allowing the
distribution of the workload among multiple cores/computers running in parallel.
The Python ecosystem already provides excellent libraries for this kind of
analysis. But merely having a good library for doing NLP, is not enough. The
most interesting problems in NLP involve large volumes of data
which can be neither stored nor processed on a single PC. PyPLN integrates a
number of open-source libraries and applications into a distributed pipeline for
linguistic analysis. 

\begin{figure}[h]
\centering
\includegraphics[width=\textwidth]{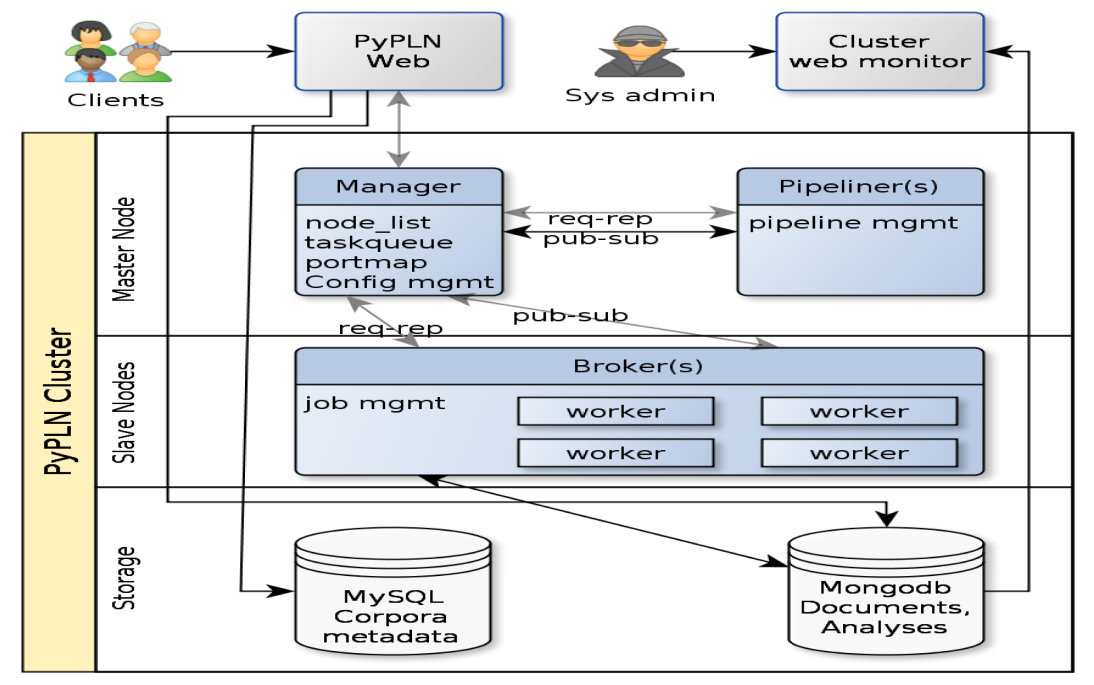}
\caption{Overview of PyPLN's architecture.}
\label{fig:arch}
\end{figure}

Being a distributed pipeline for analytical text processing, PyPLN applies the
philosophy of Unix pipeline processing to distributed computing on a cluster
environment. It provides a tool capable of handling large volumes of data while
remaining simple to deploy and program for customized analyses. PyPLN implements
distributed processing by relying on ZeroMQ for handling communication between
processes on a single SMP machine or cluster \citep{coelho2012}. Job
specifications and data are farmed to worker processes in the pipeline as JSON
messages which are balanced among the cluster. Every step of the pipeline can be
parallelized as long as it applies to more than one document. Finer grained
parallelization is also possible with the implementation of custom workers.
Workers are typically Python scripts which can, on their turn, call any
executable or library available in the system. Figure \ref{fig:arch} depicts the
general architecture of PyPLN.

A common bottleneck in the analysis of large collections of documents is disk
I/O. PyPLN relies on a NoSQL database (Mongodb) to store all the documents and
analytical results. Due to memory mapping, even a single instance of Mongodb
proves much more performant than direct file system storage. For cluster
configurations, the database can be sharded and replicated to maximize
performance and locality of the data with respect to the workers.  Any
collection
of documents in the PyPLN database is fully searchable as each new document
uploaded is immediately indexed by a high performance search engine.

\subsection{The Workers}

The worker is the basic unit of analysis in the platform. Its basic structure
was designed to be as simple as  possible, so that anyone with minimal knowledge
of PyPLN internals can write and contribute a worker to the project. Figure
\ref{fig:worker} shows the code of a worker to count the frequency of words in a
document. 

\begin{figure}[h]
\begin{lstlisting}[language=python,numbers=left]
from pypelinin import Worker


class FreqDist(Worker):
    requires = ['tokens', 'language']

    def process(self, document):
        tokens = [info.lower() for info in document['tokens']]
        frequency_distribution = {token: tokens.count(token) \
                                  for token in set(tokens)}
        fd = frequency_distribution.items()
        fd.sort(lambda x, y: cmp(y[1], x[1]))
        return {'freqdist': fd}

\end{lstlisting}
\caption{Token counter}
\label{fig:worker}
\end{figure}

Workers are Python modules with a very simple minimal structure but which can be
as complex as their tasks may require. Let's go through the required lines of
code
as can be found in the example of figure \ref{fig:worker}: It must contain an
import statement like the one shown on line 1; It must declare a Class with an
informative name, which inherits the Worker base class; This class must declare
at least one method named \emph{process}(line 7), and a global variable named
\emph{requires} (line 5). The \emph{requires} variable must be assigned a list,
containing analytical results\footnote{These may have been calculated before,
in which case they are simply read from the database.} on which this worker
depends. In this case, the
\emph{FreqDist} worker depends on the text having been tokenized before and its
language detected. The \emph{process} method takes one argument (a Python
dictionary that represents the JSON document stored in the database) and
returns a dictionary with at least one key/value pair where the key corresponds
to the name of the result provided.

Analyses which are fundamental and/or pre-requisites for other analyses are
persisted in the document database to be easily accessible for reuse.

Typically most worker implementations wraps code from an existing NLP library
such as NLTK \citep{bird2006nltk} or external executable such as
\emph{html2text}. The wrapping ensures that the calling of the external code
conforms with the requirements of the platform and can be dispatched to other
machines of the cluster to be executed in parallel.

\subsection{The Pipeline}
As mentioned before, distributed processing is implemented in Python using the
metaphor of a pipeline. Pipelines come in two basic flavors in PyPLN. The first
type are built-in ones which consist of basic preprocessing which has to be
applied to every new document added to the system (figure). The second type are
custom pipelines which can be specified by the user and submitted to the system
via the API or,  in the future, via the web interface. Figure \ref{fig:pipeline}
show the basic syntax for defining a pipeline. A pipeline is defined as a
simple Python dictionary where every key:value pair represents a pipe.
\begin{figure}
\begin{lstlisting}[language=python,numbers=left]
 pipeline = Pipeline({Job('WorkerName1'): Job('WorkerName2'),
                      Job('WorkerName2'): Job('WorkerName3')},
                    data={'foo': 'bar'})
\end{lstlisting}
 \caption{Defining a new pipeline.}
 \label{fig:pipeline}
\end{figure}

Pipelines are created and managed by the pipelinin'
library\footnote{\url{github.com/NAMD/pypelinin}}, which is part of the PyPLN
distribution. By
providing means of defining analytical pipelines, PyPLN makes it easy to store
complex analytical workflows in very high-level description language, which can
be parsed and executed by the system with maximal performance. Using the API
programmatically is another way to achieve the same results but with a little
less scalability.

\subsection{The Web Interface}

\begin{figure}[h]
\centering
\includegraphics[width=\textwidth]{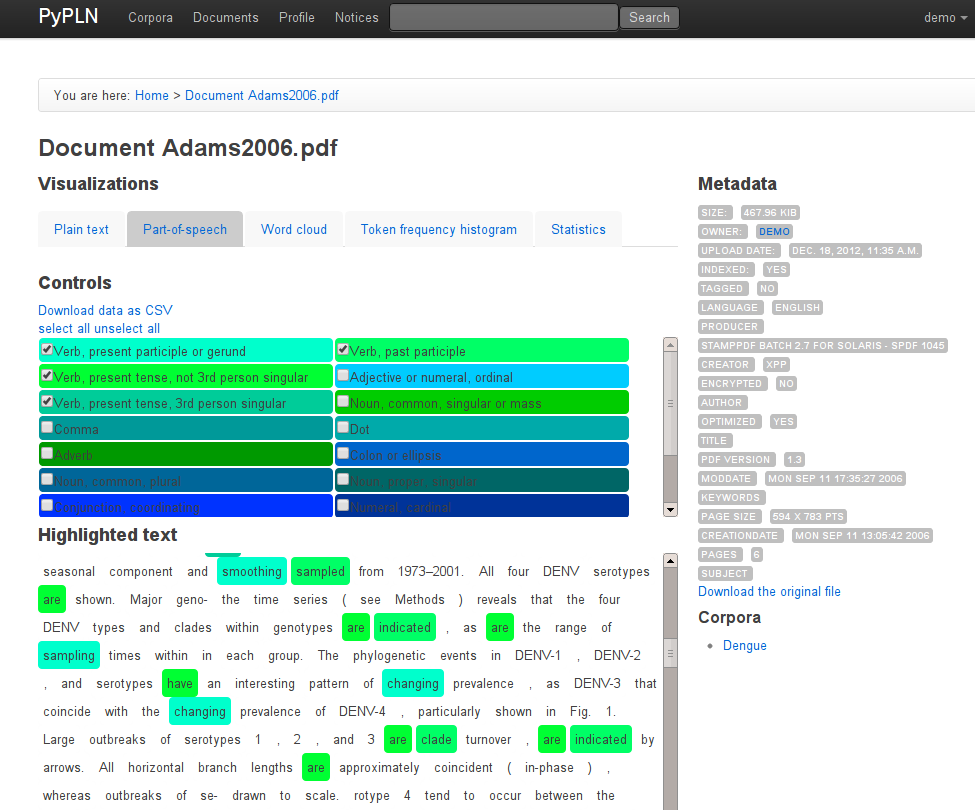}
\caption{Part-of-speech tagging analysis of PyPLN. Note the various verbal forms
highlighted in the text.}
\label{fig:postag}
\end{figure}

PyPLN offers a web-based interface for interactive text
analysis\footnote{\url{demo.pypln.org}} in which one can create and manage
corpora, perform full
text searches and visualize the analyses performed. Figure \ref{fig:postag}
shows one aspect of the web interface displaying the part-of-speech tagging of
a document.

Currently the web interface of PyPLN is the easiest way to interact with the
platform. Either through our free public server or on the user's own
infrastructure. The installation and deploy of PyPLN's backend and web
front-end is fully documented \footnote{\url{pypln.org/docs/}} and requires no
special
hardware, just a single computer running a recent version of Gnu/Linux (for a
minimal install).

When a document is uploaded via the web interface it will pass through the
workers defined in the pipeline. After the analyses are done, the user has
access to interactive visualizations on the web interface, along with links for
the data in machine readable formats. Besides being an easy to use tool for
interactive text processing, the web interface is also meant to be a showcase
of the platform's capabilities.

All the functionality provided by the web interface is also available as a REST
API, which can be used to interact with PyPLN using your own scripts. This API
is also the only way to communicate with the backend, so all connections go
through the authentication and authorization system.

\section{Final Remarks}
In this paper, we have provided a general description of the PyPLN platform,
developed to make it possible to process large quantities of unstructured
textual data. PyPLN's goal of integrating a very comprehensive array of NLP
tools into an easy to use web-based platform, is still at least one year
away. But this goal will be approached incrementally and many new features
are already scheduled for the next releases.

Another goal of the PyPLN project is to contribute linguistic knowledge back
to the academic community, in the form of annotated corpora, thesauri, and
other linguistic constructions produced by the platform's analytical engine.

PyPLN is an open-source project and welcomes the participation of anyone which
would like to see such a platform, as proposed in this paper, grow and remain
available without cost to the community at large.

Despite being on an early stage of development, PyPLN is already very usable
and has been employed to help in computational linguistic studies.
\citep{coelho2012a, coelho2012b}


\acks{We would like to acknowledge support for this project
from Funda\c{c}\~ao Getulio Vargas). }


\newpage


\vskip 0.2in
\bibliography{refs}

\end{document}